\documentclass[sn-mathphys]{sn-jnl}

\usepackage{soul}

\usepackage{academicons}
\usepackage{xcolor}

\newcommand{\orcid}[1]{\href{https://orcid.org/#1}{\textcolor[HTML]{A6CE39}{\aiOrcid}}}

\jyear{2021}%

\theoremstyle{thmstyleone}%
\theoremstyle{thmstyletwo}%

\theoremstyle{thmstylethree}%

\begin{document}

\title[A Data-driven Approach to Neural Architecture Search Initialization]{A Data-driven Approach to Neural Architecture Search Initialization}

\author[1,2]{\fnm{Kalifou Ren\'e} \sur{Traor\'e}}\email{kalifou.traore@dlr}

\author[2]{\fnm{Andr\'es} \sur{Camero}}\email{andres.camerounzueta@dlr}

\author*[1,2]{\fnm{Xiao Xiang} \sur{Zhu}}\email{xiaoxiang.zhu@dlr}

\affil*[1]{\orgdiv{Data Science in Earth Observation}, \orgname{Technical University of Munich}, \orgaddress{\street{Arcisstrasse 21}, \city{Munich}, \postcode{80333}, \state{Bavaria}, \country{Germany}}}

\affil*[2]{\orgdiv{Remote Sensing Institute}, \orgname{German Aerospace Center (DLR)}, \orgaddress{\street{ Münchener Strasse 20}, \city{Weßling}, \postcode{82234}, \state{Bavaria}, \country{Germany}}}

\abstract{

Algorithmic design in neural architecture search (NAS) has received a lot of attention, aiming to improve performance and reduce computational cost. Despite the great advances made, few authors have proposed to tailor initialization techniques for NAS. However, literature shows that a good initial set of solutions facilitate finding the optima.
Therefore, in this study, we propose a data-driven technique to initialize a population-based NAS algorithm.
Particularly, we proposed a two-step methodology. First, we perform a calibrated clustering analysis of the search space, and second, we extract the centroids and use them to initialize a NAS algorithm.
We benchmark our proposed approach against random and Latin hypercube sampling initialization using three population-based algorithms, namely a genetic algorithm, evolutionary algorithm, and aging evolution, on CIFAR-10. More specifically, we use NAS-Bench-101 to leverage the availability of NAS benchmarks.
The results show that compared to random and Latin hypercube sampling, the 
proposed initialization technique enables achieving significant long-term improvements for two of the search baselines, and sometimes in various search scenarios (various training budget).
Moreover, we analyse the distributions of solutions obtained and find that 
that the population provided by the data-driven initialization technique
enables retrieving local optima (maxima) of high fitness and similar configurations.  
}




\maketitle

\section{Introduction}

Deep learning has successfully been applied to a wide variety of problems, showing (in many cases) \emph{super human} performance~\cite{haykin2009neural,LeCun2015}. However, this success has been followed by an increasing complexity of the deep learning models, that in most cases are manually designed~\cite{elsken2019neural}. State-of-the-art \emph{deep neural networks} (DNNs) may have several million parameters, thus automating the design of DNNs is the \emph{logical} next step.

Neural architecture search (NAS) is the process of automating architecture engineering~\cite{Ojha2017,elsken2019neural}. Great advances have been made in this matter, but in practice NAS has not been adopted yet~\cite{Ojha2017,elsken2019neural}. From an optimization point of view, NAS is a challenging task. It requires dealing with huge search spaces (the deeper the model, the bigger the search space), mixed type solutions (i.e., a combination of integer, discrete, and real values to represent the model), and solutions that are \emph{expensive} to evaluate (i.e., training a DNN on a large data set may take several days).

Researchers are working to alleviate NAS challenges. Several (optimization) approaches have been tailored to design neural networks~\cite{DARTS,AgingEvol2019Real,camero2021bayesian}, to speed up model evaluation~\cite{Domhan2015,Camero2018lowcost}, and a lot of effort have been put to design NAS search spaces~\cite{AgingEvol2019Real,DARTS}. Recently, some authors have released performance evaluation databases~\cite{pmlr-v97-ying19a,dong2020nasbench201}, aiming to democratize NAS, i.e., \emph{everyone} can test a NAS algorithm regardless of having powerful computational resources, and to improve the reproducibility~\cite{elsken2019neural}. 

Despite the great advances made so far, there remain open questions. For example, many NAS approaches can be categorized as population-based algorithms~\cite{Ojha2017,elsken2019neural}. However, little attention has been drawn to the initialization of the population. Considering all the available resources, including NAS databases, we propose to address the following question: Can we improve the performance of a population-based NAS algorithm by initializing its population with a data-driven approach? 
To this problem, we propose a two-step approach.
First, a tailored clustering analysis of a target search space is performed. Second, after obtaining satisfying quantitative clustering results, the centroids are extracted and used to initialize a population based NAS algorithm.

To validate our proposal we selected three population-based NAS algorithms: an evolutionary algorithm (EA)~\cite{back1996evolutionary}, a genetic algorithm (GA)~\cite{holland1962outline}, and aging evolution (AE)~\cite{AgingEvol2019Real}, and we have benchmarked our proposed initialization technique on NAS-bench-101~\cite{pmlr-v97-ying19a} against the most popular initialization methods (random initialization and Latin hypercube sampling). 
The results show that centroids extracted using \emph{Bayesian Gaussian Mixture of models} (BGM) for clustering are a promising approach to 
initialize the population.
Particularly, our approach used with GA shows significant long-term improvements (after 2000 iterations, in test) over random initialization and Latin hypercube sampling.
When used with EA, a faster convergence (in validation) and a significant long-term improvement
over random initialization and Latin hypercube sampling is observed. 
Additional investigations on the distributions of the solutions found by the algorithms suggest 
that centroids enable retrieving local optima (maxima) of high fitness and similar configurations.

The remainder of this article is organized as follows:
The following section introduces NAS and briefly summarizes the state-of-the-art of initialization techniques. 
Section~\ref{sec:methodology} describes the proposed methodology.
Section~\ref{sec:experimental-setup} introduces the experimental setup. Section ~\ref{sec:results} presents the results.
Finally, Section~\ref{sec:conclusion} outlines the conclusions and proposes future work.

\section{Related Work}\label{sec:related-work}

In this section, we summarize some of the most relevant works related to our proposal. 
First, we introduce NAS and highlight the state-of-the art, with a special emphasis on metaheuristic approaches.
Second, we outline the population initialization problem.

\subsection{Neural Architecture Search}

NAS is the process of automating architecture engineering~\cite{elsken2019neural}. Currently, it is considered to be a subfield of AutoML~\cite{hutter2019automated}. However, its roots can be tracked to the late 1980s, where the use of \emph{evolutionary computation} was explored to design and train neural networks~\cite{engel1988fnn-sa,montana1989fnn-ga,alba1993genetic,alba1993full,yao1993review}. These ideas gather together under the \textit{neuroevolution} concept, and in the 2000s gained popularity thanks to the \emph{NeuroEvolution of Augmenting Topologies} (NEAT) method~\cite{stanley2002evolving}, a \emph{genetic algorithm} (GA) that increasingly evolves complex neural network topologies and weights. Later, due to apparition of deep learning, the neuroevolution research started to attract attention again~\cite{Ojha2017,elsken2019neural}. 

From the optimization (algorithm) point of view, many approaches can be found in the neuroevolution literature, ranging from evolutionary algorithm (EA)~\cite{camero2019specialized}, GA~\cite{zhining2015genetic}, \emph{harmony search}~\cite{rosa2015cnnhs} and \emph{mixed integer parallel efficient global optimization} technique~\cite{van2019automatic}, to Bayesian optimization~\cite{camero2021bayesian}. And also, from the point of view of the neural network architecture, e.g., recurrent neural network~\cite{ororbia2019investigating}, convolutional neural network~\cite{miikkulainen2019evolving} and generative adversarial networks~\cite{wang2019evolutionary}.

On the other hand, in the past few years, a \emph{new branch} of NAS approaches emerged based on \emph{continuous optimization} (e.g., DARTS~\cite{DARTS}). Particularly, these approaches search over a large graph of overlapping configurations (i.e., the \emph{Super-Net}) using a gradient-based approach.
These recent improvement result in large speedups in terms of search time~\cite{elsken2019neural} but also in some case in a lack of robustness and interpretability~\cite{Yang2020NasEvalHard}.

Despite the NAS approach used, literature stress the importance of optimizing the architecture of a deep neural network (given a particular problem). The main challenges of NAS are three-fold: First, the number of the parameters increases in proportion to the number of layers, thus the search space is huge. Second, the search space is (usually) a mix of categorical (e.g., the type of operation, the activation functions, ...), real (e.g., the weights), and integer (e.g., the number of hidden layers, the number of neurons per layer, ...) or discrete (e.g., the adjacency matrix) values, resulting in a complicated problem, i.e., each parameter type require a different optimization approach. Third, the evaluation of an architecture is extremely resources and time-consuming. Therefore, NAS problems fall into the family of expensive optimization problems. 

To cope with the latter problem, i.e., the evaluation cost, and aiming to improve reproducibility, lots of effort have been made to provide open-source benchmarks for NAS. Several areas of applied \emph{machine learning} have been included in these benchmarks, including \emph{computer vision}~\cite{ying2019nasbench101,dong2020nasbench201,zela2020nasbench1shot1,siems2020nasbench301} and \emph{natural language processing}~\cite{Klyuchnikov2020NasBenchNLP}, among others~\cite{elsken2019neural}. Also, some authors have explored techniques to speed up the performance evaluation~\cite{Domhan2015,Camero2018lowcost}.
The mixed search space problem has been faced from multiple perspectives, ranging from tailored encoding~\cite{durr2006neuroevolution,ning2020generic}, and specific operations~\cite{ororbia2019investigating}, to mixed (hybrid) approaches~\cite{chu2021fast}.

Finally, to tackle the problems that arise due to the size of the search space (first challenge), several authors have invested time tailoring the design of the search space~\cite{AgingEvol2019Real,DARTS}, providing tools to assess its \emph{quality}~\cite{FLA_NAS_GNNS_Nunes}\cite{traore2021fitness}, and proposing techniques to adapt the search space~\cite{zhang2021adaptive,camero2021bayesian}, among others. Despite all the advances made in this regard, the initialization of the NAS algorithms (especially the population-based ones) has not received much attention. However, it is important to remark that starting from a set of \emph{good} solutions is key to solve large-scale optimization problems using a population of finite size~\cite{maaranen2004quasi}.

\subsection{Population Initialization Techniques}

All population-based metaheuristic algorithms share a common step: population initialization. The goal of this initialization is to provide a first set of solutions, that (normally) will be improved in an iterative way until the termination criteria is met. How \emph{good} (or \emph{bad}) is the initial population facilitate (or prevent) finding the optima~\cite{rahnamayan2007quasi,clerc2008initialisations,maaranen2004quasi}, and this is more serious for large-scale optimization problems using a population of finite size~\cite{maaranen2004quasi}, which is the most common case, specially for expensive to evaluate problems (like NAS). Therefore, the greater the search space (given a limited population size), the smaller the chance to cover promising regions of the search space~\cite{helwig2008theoretical}.

In the past few decades, some authors have started to propose initialization techniques aiming to boost the performance of population-based metaheuristic algorithms (mainly EA)~\cite{Kazimipour2014Review_EA_Init,kazimipour2013initialization}. Great advances have been made, for example, \cite{kimura2005genetic} shows that initialization can increase the probability of finding global optima, another study shows that stability can be improved~\cite{morrison2003dispersion}, in \cite{ma2012impact} it is show that the solution quality is related to the initialization, among many others~\cite{Kazimipour2014Review_EA_Init,kazimipour2013initialization}. 

However, in black-box optimization problems, such as NAS, it is not possible to determine beforehand what is a good and bad solution. Therefore, not all initialization techniques are suitable for NAS. Moreover, few practical rules of thumbs are provided in the literature to choose an appropriate initialization technique. Thus, from a practitioner perspective, it is unclear how to choose the right initialization technique~\cite{Kazimipour2014Review_EA_Init}. 

Considering all these limitations, most practical NAS implementations rely on (quasi)random~\cite{kimura2005genetic} initialization or Latin Hypercube Sampling (LHS)~\cite{poles2009effect,camero2021bayesian,mousavirad2019tackling,medeiros2019latin}. Therefore, in this study we propose to tackle the population initialization problem for NAS.

\section{A data-driven approach to initializing a NAS search strategy} \label{sec:methodology}

This section introduces the proposed approach of Cluster Analysis for enhancing the performances of a NAS algorithm. 
First, we describe the overall pipeline of the methodology.
Second, we detail the feature engineering essential to the analysis.

\subsection{Pipeline}

This study aims at leveraging the knowledge about a Search Space to help improve 
the performances of a Search Strategy.
In particular, it sets out to answer the following question:
can we improve the convergence of a population-based NAS algorithm 
by initializing it with a data-driven approach?

\begin{figure}[ht]
  \includegraphics[width=0.99\linewidth]{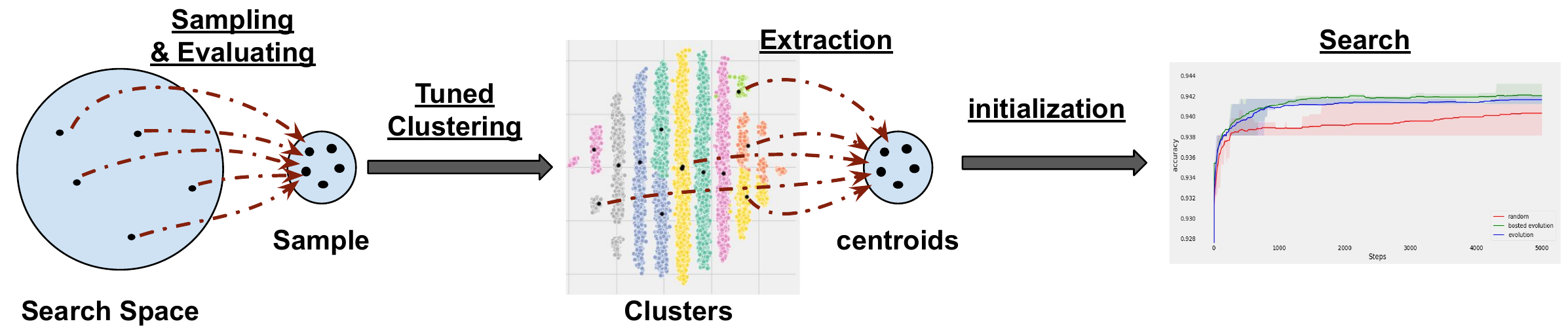}\hfill
  \caption{ The pipeline of the proposed data-driven approach to NAS initialization. }
  \label{fig:pipeline}
\end{figure}

To tackle this problem, we propose an approach consisting of two steps, depicted in Figure~\ref{fig:pipeline}.
First, we perform a performance-based clustering analysis of the Search Space.
Given the search space and a machine learning task, we sample a set of $N$ architectures. 
Each architecture is trained, evaluated, and encoded using the procedures described in Section~\ref{subsec:encoding}.
As the feature vector consists of an architecture representation and its performance in test, 
the resulting clusters should relate to specific behaviors (performances) on the learned task. 
Moreover, processing high dimensional and sparse data can sometimes be uneasy, therefore we propose to facilitate the clustering by reducing the dimension of input features.
With the reduced samples, we proceed with a clustering analysis composed itself of a sequence of sub-steps.
This sequence is as follows: 
we reduce the dimension of the samples, 
perform the clustering and assessing qualitatively and quantitatively its results.
Besides, calibrating the proper number of clusters play an important role 
in the results retrieved. 
Thus, we seek to identify values to this hyperparameter providing satisfying results.


In the second step of the proposed methodology, 
we extract the centroids obtained in the clustering, 
initialize a population-based algorithm, 
and assess it performances.   

To summarize, this approach to search initialization is: 
\begin{itemize}
    \item \textit{Composite}: It is a multistep initialization procedure relying on sampling a search space,
    clustering it, and initializing an algorithm with the centroids extracted.
    \item \textit{Generic}: It is not application-specific, in fact the clustering could be done on any type of search space given an encoding including a solution representation and its fitness evaluation.
    \item \textit{Deterministic or stochastic}: The stochasticity of the procedure depends on the stochasticity of the tool selected for clustering.
\end{itemize}

Also, it is important to note that the clusters may be analyzed to obtain insights regarding the \emph{archetypes} (i.e., the representative architectures), including the most frequent operations and the connection between the operations (i.e., the edges in the graph).

\subsection{Feature representation} \label{subsec:encoding}

To best take advantage of information about the search space when clustering, 
we first introduce a minimal feature engineering.

As we look to uncover models and structures relevant to NAS algorithms via clustering, 
we seek a feature representation encoding an architecture as well as its performances.
As in~\cite{pmlr-v97-ying19a}, we consider neural architectures identified by an elementary component repeated in blocks, a feed-forward cell.
This cell is a directed acyclic graph (DAG), with a maximum number of operations (nodes),
a maximum number of transformations (edges) and a fixed set of possible operations (e.g., max pool, convolution 3x3) labeling each node.
A cell is in practice represented a list of selected operations and an adjacency matrix of variable size.

Therefore, we construct two versions of clustering feature representation, both in the form of vectors.
The first one (\textbf{Original}, Short Encoding) consists in concatenating for each model, its adjacency matrix,
the list of operations, and the list of performances in test for all available training duration $\{t_0, t_1, t_2, t_3\}$.
Note that this is a variable length feature representation due to the nature of the adjacency matrix.

Alternatively, the second representation (\textbf{Binary}, Long Encoding) corresponds to the expanded adjacency matrix, i.e., the matrix that consider all possible operations (according to the constrains of the search space). This is a fixed length encoding. 
Moreover, for both encoding, the vector form of the adjacency matrix is obtained 
by a flattening in row-major fashion (C-style).

\section{Experimental Setup}\label{sec:experimental-setup}

The experiments performed aim to validate that the initialization of a population-based NAS Algorithm can benefit from models identified via Clustering Analysis of a Search Space.
In this Section, first, we introduce the problem used to validate our proposal. Second, we present the parameters used for performing the experiments on clustering.
Third, we detail the performance metrics used to assess the quality of the clustering. 
Fourth, we describe the three (3) baseline algorithms that we aim to initialize.

\subsection{NASBench-101} \label{sec:nb101-dataset}

NASBench-101 is a database of neural network architectures and their performance evaluated on the data set of CIFAR-10. It contains $N=450K$ unique architectures~\cite{pmlr-v97-ying19a}.
Indeed, to tackle the given machine learning task of CIFAR-10,
all contained models use of a classical image classification structure similar to ResNet.
Indeed, the backbone of a model contains a head, a body and a tail.
Its body is made by alternating three (3) times a block with a down-sampling module.
Each block is obtained by repeating three (3) times a module called 'cell'. 
A cell is a computational unit that can be represented by a DAG.
It consists of an input node, an output node and intermediate nodes representing operations (convolution 3x3, convolution 1x1, maxpool 3x3), and connections indicating features being transformed. Therefore, each architecture differs by its cell. 
In practice, the DAG of a model is encoded by an adjacency matrix and a list of operations labelling the associated nodes. The constraints on such DAG are the following: There can be at most $N=7$ nodes and $E=9$ edges in a cell. 

Moreover, all models were trained for 108 epochs using the same experimental setting (i.e., learning rate, etc.), but performance evaluations in training, validation and test were also provided after 4, 12 and 36 epochs.

\subsection{Hyperparameters for Clustering}

The clustering experiments were done with a set of randomly sampled models.
The size of the sample is identified in Section~\ref{subsec:sample-size}.
The considered clustering algorithms are \emph{k}-means, DBSCAN, BIRCH, spectral clustering, and a BGM. All 
were obtained from the latest version (0.24.1) of the \textit{scikit-learn} library~\cite{scikit-learn}. 
Table~\ref{table:hyperparam-clustering} shows the hyperparameters selected for each method, including
the maximum number of iterations (\emph{Max iter}), the number of samples used at initialization (\emph{N init}), 
and other algorithm-specific parameters (\emph{Other}).
Note that they are either default (\texttt{NA}) or 
slightly modified to provide satisfying 
clustering performances. 

\begin{table}[ht]
 \centering 
 \scriptsize
\begin{tabular}{lrrr}\hline 
\textit{Method} & \textit{Max iter} & \textit{N init} &  \textit{Other} \\\hline 
KMEANS & 500 & 50 & kmeans++ init \\
DBSCAN & 500 &  200 & eps=0.30 \\
BIRCH & 500 & \texttt{NA} & threshold=0.12\\
SPECTRAL & 500 & \texttt{NA} & \texttt{NA} \\
BGM  & 500 & \texttt{NA} & Dirichlet weight distribution, full co-variance\\
\hline  
\end{tabular}
\caption{Hyperparameters of the clustering algorithms.}
\label{table:hyperparam-clustering}
\end{table}

\subsection{Clustering performance evaluation}\label{subsec:clustering-metrics}

Moreover, we use various ways of assessing the quality of the results for each step of the approach. 
Regarding step one (1), we propose to measure the clustering performance using the following three (3) standard metrics: Silhouette coefficient~\cite{SilhouetteCoefficient}, Calinski-Harabasz~\cite{CalinskyHarabaszIndex} and Davies-Bouldin index~\cite{DaviesBouldinIndex}.
These metrics inform on how well separated and dense are the resulting clusters.
They all apply in the context of clustering with missing labels, which is relevant, as we 
seek to investigate relevant clusters and features for NAS algorithms without prior assumptions. 
The Silhouette coefficient is a metric comprised between -1 and +1, with higher values associated to more dense and separated clusters
The Calinski-Harabasz index also rates a better defined clusters with higher values.
Similarly, the Davies-Bouldin index, measures a \emph{similarity} between clusters, providing smaller values for better clustering.
%
%
Additionally, we propose to corroborate the quantitative assessment with a qualitative analysis (visual inspection) for validation before step two.

Regarding step two (2), we assess the quality of the centroid-based initialization using the performance of the algorithm (best obtained accuracy in test). More importantly, we compare the performance against initializing the same algorithm with random and LHS initialization. Also, as a sanity check, we compare the results against \emph{random search}.

\subsection{Baseline Algorithms}\label{sec:baseline-algos}

To benchmark the performance of the different initialization techniques, we propose to use three (3) population-based algorithms. Particularly, we implemented a Genetic Algorithm (GA), an Evolutionary Algorithm (EA), and \emph{Aging Evolution} (AE)~\cite{AgingEvol2019Real}, a popular EA-based algorithm specifically designed for NAS on CV problems.

\subsubsection{Genetic Algorithm}\label{sec:ga}

A GA is a population-based meta-heuristic algorithm inspired by natural evolution~\cite{holland1962outline}. At a glance, a \textit{population} of \textit{individuals} (a.k.a. solution) is evolved using selection, crossover, mutation, and replacement operations. Particularly, we used the implementation available in the latest version (1.3.1) of DEAP library~\cite{DEAP_JMLR2012}. Algorithm~\ref{algo:ga} presents a high-level view of the implemented GA.

\begin{algorithm}[!h]
\SetAlgoLined
\DontPrintSemicolon
\textbf{input:} The size of the population \textit{pop\_size}, the crossover probability \textit{cx\_p}, mutation probabilities \textit{mut\_p} and \textit{mut\_i}, and the maximum number of evaluations \textit{max\_evaluations}. Alongside with \textit{train}, \textit{validation}, and \textit{test} data sets. \;
$ \textsc{solution} \gets \emptyset $\;
$ \textsc{population} \gets \operatorname{Initialize}(\textit{pop\_size})$ \;
$ \operatorname{Evaluate}(\textsc{population}, \textit{train}, \textit{validation}) $\;
$ \textsc{evaluations} \gets \textit{pop\_size} $\;
$ \textsc{solution} \gets \operatorname{Best(\textsc{solution}, \textsc{population})} $\;
\While{$ \textsc{evaluations} \leq \textit{max\_evaluations} $}{
    $ \textsc{offspring} \gets \emptyset $\;
    \For{$j \gets 1 \,\, \text{to} \,\, \textit{pop\_size}$}{
        $ \textsc{parent\_1} \gets \operatorname{BinaryTournament(\textsc{population}) }$\;
        $ \textsc{parent\_2} \gets \operatorname{BinaryTournament(\textsc{population}) }$\;
        $ \textsc{child} \gets \operatorname{SinglePointCrossover}(\textsc{parent\_1}, \textsc{parent\_2}, \textit{cx\_p}) $\;
        $ \textsc{offspring} \gets \textsc{offspring} + \textsc{child} $
    }
    $ \textsc{offspring} \gets \operatorname{Mutate}(\textsc{offspring}, \textit{mut\_p}, \textit{mut\_i}) $\;
    $ \operatorname{Evaluate}(\textsc{offspring}, \textit{train}, \textit{validation}) $\;
    $ \textsc{population} \gets \textsc{offspring} $\;
    $ \textsc{solution} \gets \operatorname{Best(\textsc{solution}, \textsc{population})} $\;
    $ \textsc{evaluations} \gets \textsc{evaluations} + \textit{pop\_size} $\;
}
$ \textsc{performance} \gets \operatorname{Evaluate}(\textsc{solution}, \textit{train}, \textit{test}) $\;
$ \textbf{return}\,\, \textsc{solution}, \textsc{performance} $\;
\caption{Genetic Algorithm}
\label{algo:ga}
\end{algorithm}

Particularly, an individual encodes a neural network architecture (in the given search space) by a mix of binary entries, that represent the adjacency matrix of the architecture, and categorical values, that correspond to the operations on the edges of the adjacency matrix. Please refer to Section~\ref{sec:nb101-dataset} for more details.

An initial \textit{population} of size \textit{pop\_size} is initialized by the function \texttt{Initialize($\cdot$)}. Particularly, we define three variations to initialize the population: Random initialization, LHS, and our proposed method (Section~\ref{sec:methodology}). An \textit{individual} is evaluated by the function \texttt{Evaluate($\mathcal{D}_1$, $\mathcal{D}_2$)}. The decoded architecture is trained using SGD on $\mathcal{D}_1$ data set, and evaluated (accuracy) on $\mathcal{D}_2$ data set (a.k.a., the fitness). Then, the best \textit{solution} (i.e., the one with the highest accuracy) of the \textit{population} is selected by the \texttt{Best($\cdot$)} function.

Then, the evolution takes place. First, an \textit{offspring} of size \textit{pop\_size} is created. More specifically, each \textit{offspring} individual is created by a single point crossover operation \texttt{SinglePointCrossover($\mathcal{P}_1$, $\mathcal{P}_2$, \textit{cx\_p})} with probability \textit{cx\_p}, where $\mathcal{P}_i$ is selected using a binary tournament operation \texttt{BinaryTournament($\cdot$)}. Note that with probability 1 - \textit{cx\_p} one of the parents $\mathcal{P}_i$ is returned unmodified.
Later, the \textit{offspring} is mutated with probability \textit{mut\_p} by the function \texttt{Mutate($\cdot$)}. If mutated, each position is mutated using bit-flip (for the binary entries) or round-robin (for the categorical values) with probability \textit{mut\_i}.
The \textit{offspring} is evaluated using \texttt{Evaluate($\cdot$)}, and the current \textit{population} is replaced by the \textit{offspring}.
Finally, the best \textit{solution} is updated, i.e., if the fitness of the best individual in the \textit{population} is higher than the current best \textit{solution}, then the best individual become the best \textit{solution}.
Once the number of \textit{max\_evaluations} is reached, the best solution is evaluated using the \textit{test} data set.

\subsubsection{($\mu + \lambda$) Evolutionary Algorithm}

The ($\mu+\lambda$)EA~\cite{back1996evolutionary}, a generic population-based metaheuristic algorithm, evolves a population of $\mu$ individuals by creating $\lambda$ offspring. Then, both the original population and the offspring are combined, and the best $\mu$ individuals replace the population. Algorithm~\ref{algo:mulambda} presents a high-level view of the ($\mu+\lambda$)EA basic implementation provided by the latest version (1.3.1) of DEAP library~\cite{DEAP_JMLR2012}.

\begin{algorithm}[!h]
\SetAlgoLined
\DontPrintSemicolon
\textbf{input:} The size of the population $\mu$, the size of the offspring $\lambda$, mutation probabilities \textit{mut\_p} and \textit{mut\_i}, and the maximum number of evaluations \textit{max\_evaluations}. Alongside with \textit{train}, \textit{validation}, and \textit{test} data sets. \;
$ \textsc{population} \gets \operatorname{Initialize}(\mu)$ \;
$ \operatorname{Evaluate}(\textsc{population}, \textit{train}, \textit{validation}) $\;
$ \textsc{evaluations} \gets \mu $\;
\While{$ \textsc{evaluations} \leq \textit{max\_evaluations} $}{ 
    $ \textsc{offspring} \gets \operatorname{RandomSample}(\textsc{population}, \lambda) $\;
    $ \textsc{offspring} \gets \operatorname{Mutate}(\textsc{offspring}, \textit{mut\_p}, \textit{mut\_i}) $\;
    $ \operatorname{Evaluate}(\textsc{offspring}, \textit{train}, \textit{validation}) $\;
    $ \textsc{population} \gets \operatorname{RankSelection}(\textsc{population} + \textsc{offspring}, \mu) $\;
    $ \textsc{evaluations} \gets \textsc{evaluations} + \lambda $\;
}
$ \textsc{solution} \gets \operatorname{Best}(\textsc{population}) $\;
$ \textsc{performance} \gets \operatorname{Evaluate}(\textsc{solution}, \textit{train}, \textit{test}) $\;
$ \textbf{return}\,\, \textsc{solution}, \textsc{performance} $\;
\caption{$(\mu + \lambda)$ Evolutionary Algorithm}
\label{algo:mulambda}
\end{algorithm}

The \textit{population} (refer to Section~\ref{sec:ga}) of size $\mu$ is initialized using the \texttt{Initialize($\cdot$)} function. Then, the \textit{population} is evaluated using the \texttt{Evaluate($\cdot$)} function (refer to Section~\ref{sec:ga}). Then, the evolutionary process takes place. First, an \textit{offspring} of size $\lambda$ is generated by randomly sampling (with uniform probability) individuals from the \textit{population}. Following, the \textit{offspring} is mutated using the \texttt{Mutate($\cdot$)} function (refer to Section~\ref{sec:ga}), and the offspring is evaluated. In the last evolutionary step, the \textit{population} and the \textit{offspring} are combined, ranked according to their fitness, and the top $\mu$ individuals are selected by the \texttt{RankSelection($\cdot$)} to replace the current \textit{population}.

Once the number of evaluations is greater than \textit{max\_evaluations}, the best individual of the \textit{population} is selected by \texttt{Best($\cdot$)}, i.e., the \textit{solution}. Finally, the \textit{solution} is evaluated on the \textit{test} data set.

\subsubsection{Aging Evolution}

A few years ago, the \emph{Aging Evolution} (AE)~\cite{AgingEvol2019Real}, an \textit{EA}-based approach to NAS, became popular because it achieved state-of-the-art performance on classical CV benchmarks. Algorithm~\ref{algo:aging-evol} outline AE. Notice that the nomenclature does not match exactly the one proposed in~\cite{AgingEvol2019Real}, instead the algorithm presents a version that is \emph{closer} to Algorithms~\ref{algo:ga} and~\ref{algo:mulambda}. Also, we used the implementation available on NASBench-101 repository.

\begin{algorithm}[h]
\SetAlgoLined
\DontPrintSemicolon
\textbf{input:} The size of the population \textit{pop\_size}, the size of the tournament \textit{k}, and the maximum number of evaluations \textit{max\_evaluations}. Alongside with \textit{train}, \textit{validation}, and \textit{test} data sets. \;
$ \textsc{solution} \gets \emptyset $\;
$ \textsc{population} \gets \operatorname{Initialize}(\textit{pop\_size}) $\;
$ \operatorname{Evaluate}(\textsc{population}, \textit{train}, \textit{validation}) $\;
$ \textsc{solution} \gets \operatorname{Best(\textsc{solution}, \textsc{population})} $\;
$ \textsc{evaluations} \gets \textit{pop\_size} $\;
\While{$ \textsc{evaluations}  \leq \textit{max\_evaluations} $ }{ 
    $ \textsc{offspring} \gets \operatorname{Tournament}(\textsc{population}, \textit{k}) $\;
    $ \textsc{offspring} \gets \operatorname{Mutate}(\textsc{offspring}) $\;
    $ \operatorname{Evaluate}(\textsc{offspring}) $\;
    $ \textsc{solution} \gets \operatorname{Best(\textsc{solution}, \textsc{offspring})} $\;
    $ \operatorname{Enqueue}(\textsc{population}, \textsc{offspring}) $\;
    $ \operatorname{Dequeue}(\textsc{population}, \textsc{offspring}) $\;
    $ \textsc{evaluations} \gets \textsc{evaluations} + 1 $\;
}
$ \textsc{performance} \gets \operatorname{Evaluate}(\textsc{solution}, \textit{train}, \textit{test}) $\;
$ \textbf{return}\,\, \textsc{solution}, \textsc{performance} $\;
\caption{Aging Evolution}
\label{algo:aging-evol}
\end{algorithm}

AE is a steady state EA, where the \emph{oldest} individual of the population is replaced by the offspring. Particularly, the population of size \textit{pop\_size} is initialized using the function \texttt{Initialize}($\cdot$), evaluated using the function \texttt{Evaluate($\cdot$)} (we are \emph{reusing} the function defined above in this section), and the best \textit{solution} is selected from the population by the function \texttt{Best($\cdot$)}. 

Then, the evolution begins. 
First, an individual (i.e., the \textit{offspring}) is selected using a \textit{k} tournament selection. 
Second, the \textit{offspring} is mutated by a two-step process \texttt{Mutate($\cdot$)}: \emph(i) a \textit{hidden state mutation}, the connections between operations in a graph-represented solution (cell) are modified, and \emph(ii) an \textit{operation mutation}, the operation within the \textit{cell} is modified.
Then, the \textit{offspring} is evaluated, and if its performance is higher than the previous best seen \textit{solution}, the solution is replaced by the \textit{offspring} using the function \texttt{Best($\cdot$)},
In the last step of the evolution, the \emph{oldest} individual of the population (i.e., the earliest evaluated one in the population) is replaced by the \textit{offspring} using the \texttt{Enqueue($\cdot$)} (add the new one) and \texttt{Dequeue($\cdot$)} (remove the oldest one) functions.
The authors of AE claim that exists a parallel between the introduced age-based removal to a \textit{regularization} of the evolution. 

The evolution continues until the number of evaluated candidate solutions exceed the predefined budget \textit{max\_evaluations}. Finally, the best \textit{solution} of the population is returned.



\section{Results}\label{sec:results}

In this section, 
we present results on clustering for accelerating NAS algorithms.
First, we show results on selecting the proper dimension reduction tool and hyperparameters for the clustering.
Then, we show results on identifying the number of clusters providing satisfying clustering performances.
We also present results on qualitatively assessing the clusters quality for various algorithms.
Then, we provide results on improving NAS performances using a centroid-based initialization for three (3) NAS evolutionary algorithms.
Last but not least, we show results of a quantitative assessment for solutions found from the bench-marking, 
in the form of matrices of cell occurrence.

\subsection{Dimension Reduction}

To begin our experimental study, we seek to calibrate the dimension reduction of the input features.
To do so, we arbitrarily fix the number of samples to $N=10000$, in order to perform relevant experiments.

\begin{figure}[ht]
  \includegraphics[width=0.99\linewidth]{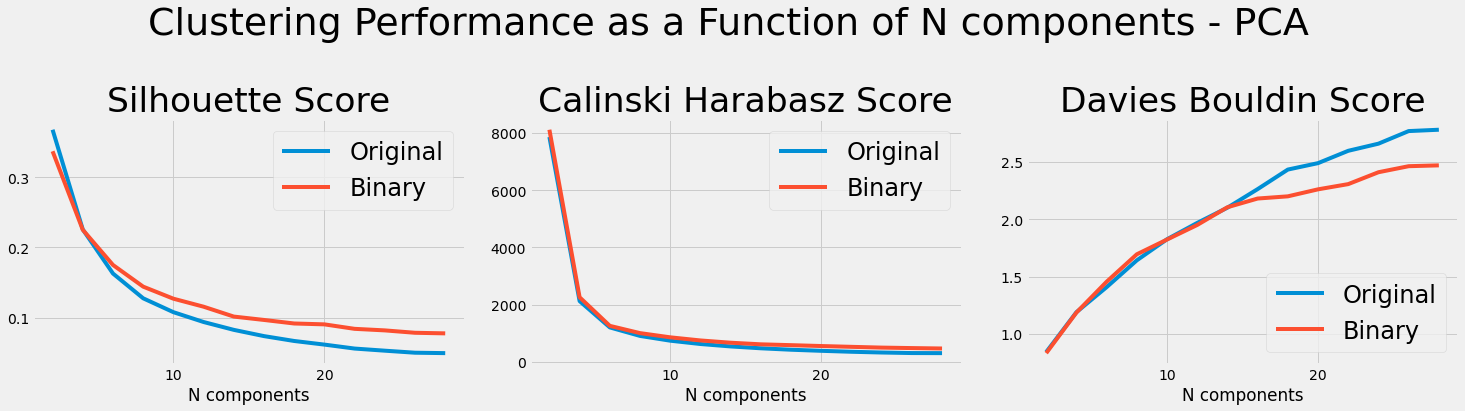}\hfill
  \caption{Input feature reduction for clustering, with an arbitrary number of clusters $N=10$. }
  \label{fig:dim-red-pca}
\end{figure}

Figure~\ref{fig:dim-red-pca} shows clustering performance as a function of the number of components of input features.
The blue and red curves display performance using respectively the Short (Original) and the Long encoding (Binary).
The dimension reduction is performed using PCA and clustering with \emph{k}-means.

Using the Short encoding, the three metrics are in favor of using a small number 
of components for input features via PCA. Indeed, the smaller the number of components the higher the Silhouette and Calinski-Harabasz scores, 
and the lower the Davies-Bouldin index, with optimal values for using two (2) components.
The same is observed when using the Long encoding.
    
Using the Long encoding yields slightly better performance than the Short encoding,
with a sensible improvement on larger number of components with PCA.

\begin{figure}[ht]
\centering
  \begin{subfigure}{\linewidth}
  \includegraphics[width=0.99\linewidth]{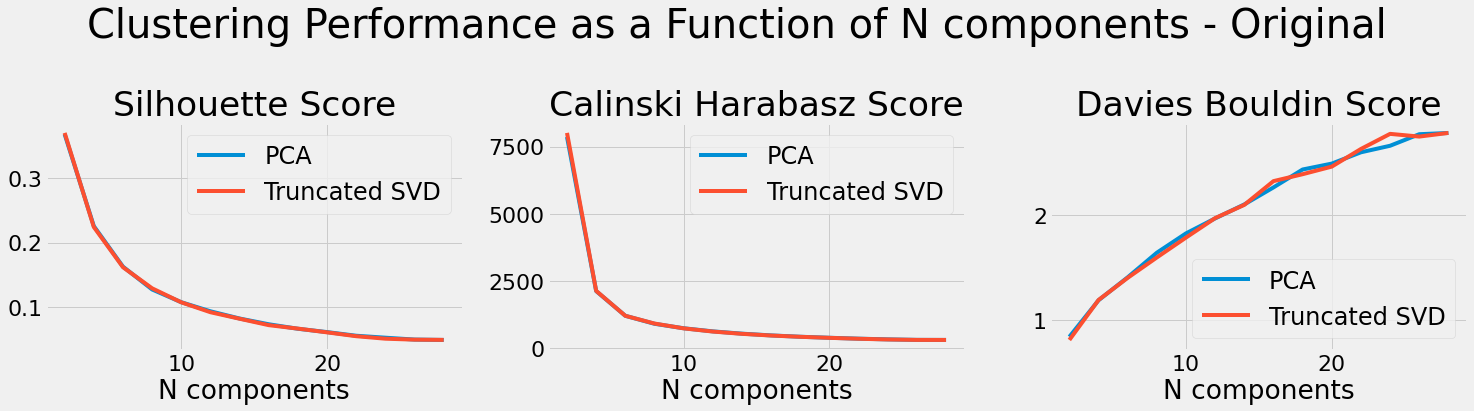}\hfill
  \caption{}
  \end{subfigure}\par\medskip
  \begin{subfigure}{\linewidth}
  \includegraphics[width=.99\linewidth]{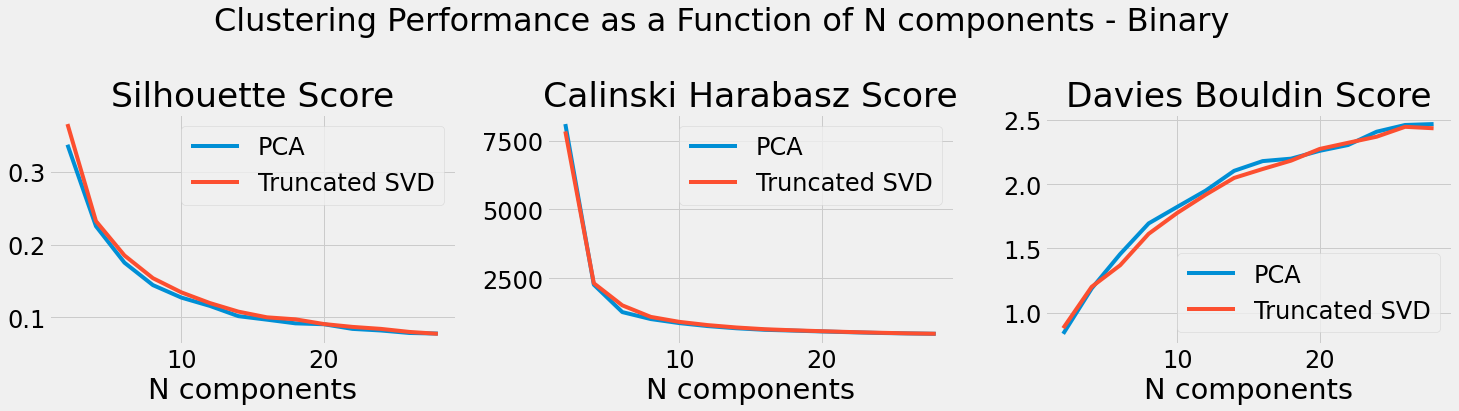}\hfill
  \caption{}
  \end{subfigure}
  \caption{Identifying the proper reduction tool for sparse data, using an arbitrary number of clusters of $N=10$.}
  \label{fig:pca-vs-truncsvd}
\end{figure}

As both encoding are rather sparse (length of up to 58, or up to 298), we study the effect of this sparsity in the dimension reduction tool.
Figure~\ref{fig:pca-vs-truncsvd} shows clustering performances as a function of the number of components of input features, for various reduction tool.
The blue and red curves display performances using respectively PCA and Truncated SVD as dimension reduction tools.
Plot (a) and (b) display results using respectively the Short and the Long encoding.
The clustering is performed with \emph{k}-means.

Trying an alternative dimensional reduction tool (Truncated SVD) more suitable for highly sparse data
does not worsen results on the Short encoding (see Figure ~\ref{fig:pca-vs-truncsvd}a). Moreover, it allows for a slight improvement over PCA when using the Long encoding (see Figure ~\ref{fig:pca-vs-truncsvd}b).

To summarize, the findings show that reducing the dimensions of the input features to 2D provides the best performances on both encoding. Using the Long encoding improves the results.
Also, using Truncated SVD shows slight improvements as it is more suitable for sparse data.
Given these findings, the following experiments are performed using Truncated SVD for a 2D reduction of input.

\subsection{Number of Samples}
\label{subsec:sample-size}

Having identified a suitable dimension reduction tool (Truncated SVD) 
and value for the number of components to reduce to (N=2), 
we now seek to find a satisfying number of samples for the clustering.

\begin{figure}[ht]
  \includegraphics[width=0.99\linewidth]{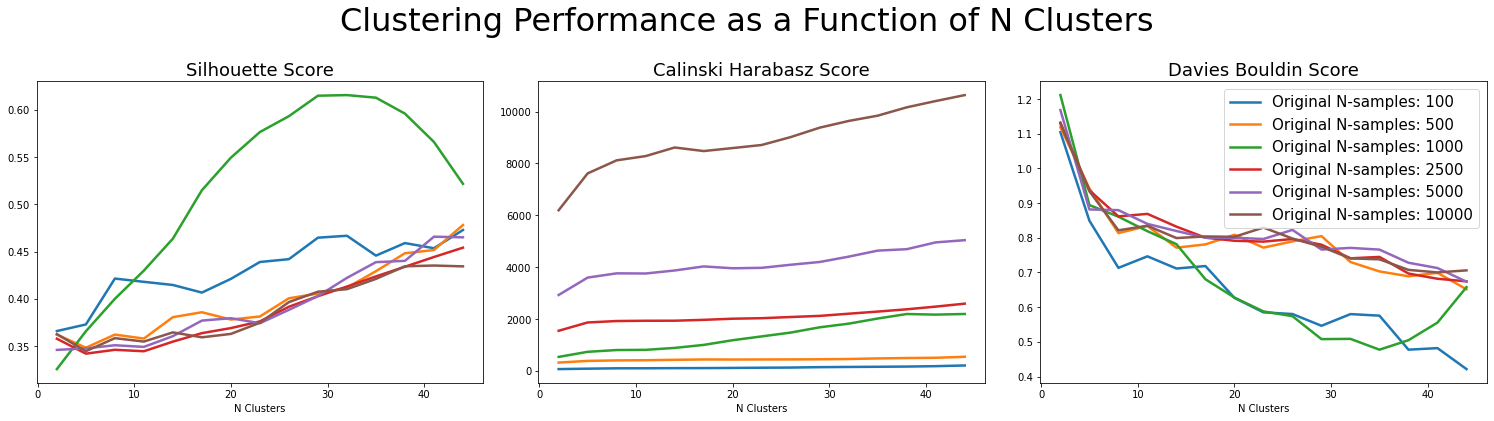}\hfill
  \caption{Identifying the proper sample size, when clustering using Truncated SVD for dimension reduction ($N=2$ components) and \emph{k}-means.}
  \label{fig:n-clusters-sample-size-short}
\end{figure}

Figure~\ref{fig:n-clusters-sample-size-short} shows clustering performances as a function of the number of clusters,
for various sample sizes.  
All were obtained when clustering with the Short encoding for feature representation. 
The blue, orange, green, red, purple and brown curves are for respectively using 100, 500, 1000, 2500,
5000 and 10000 samples.
This range of values enables us to consider small to intermediately large complexity for our proposal.

Overall, all performance metrics points towards the use of large number of clusters.
Indeed, the higher the number of clusters, the higher the Silhouette and Calinski-Harabasz scores, and the lower the Davies-Bouldin index. 
These trends are observed for all sample sizes, with similar values for Silhouette Score and the Davies-Bouldin index.
Only the Calinsky-Harabasz index discriminates towards the use of increasingly larger sample sizes.

Given these findings, we identify $N=10000$ (the largest tested value) as the sample size to use for optimal clustering in future experiments.

\subsection{Number of Clusters}

Next, we look closer into the number of clusters to use, for when clustering with various feature representations.

\begin{figure}[ht]
  \includegraphics[width=0.99\linewidth]{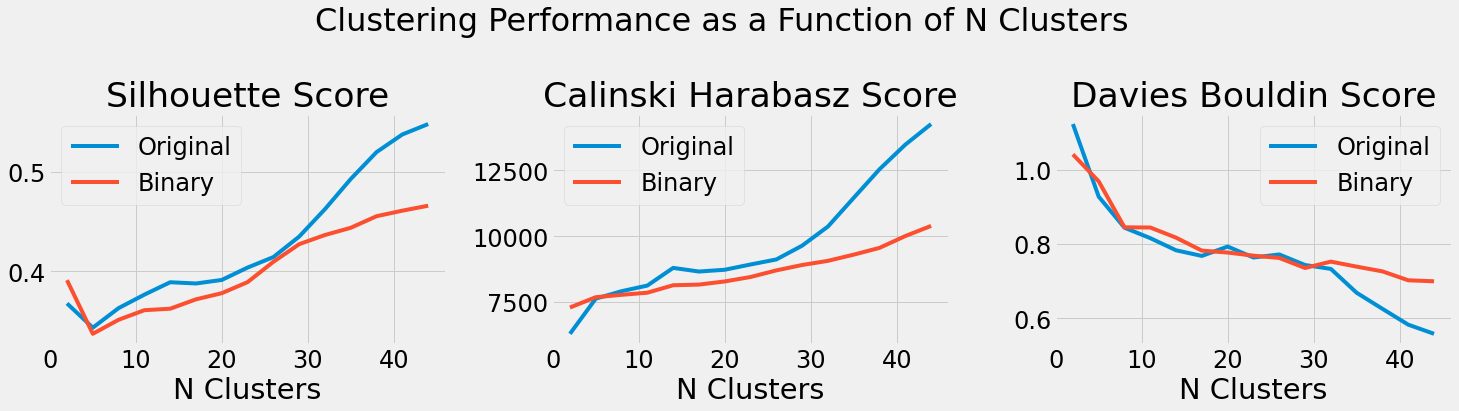}\hfill
  \caption{Identifying the proper number of clusters, using Truncated SVD for dimension reduction ($N=2$ components) and \emph{k}-means.
  }
  \label{fig:n-clusters}
\end{figure}

Figure~\ref{fig:n-clusters} shows clustering performance as a function of the number of clusters.
The blue and red curves display the performance results using respectively the Short and the Long encoding. All input features were reduced to two (2) components using Truncated SVD, 
and clustering is performed with \emph{k}-means.

Using both encoding, all performance metrics points towards the use of large number of clusters.
The higher the number of clusters, the higher the Silhouette and Calinski-Harabasz scores, and the lower the Davies-Bouldin index. 
Additionally, intermediate values around twenty (20) and twenty-seven (27) clusters respectively for the Original and Binary encoding seem to reach satisfying performance already. 

Therefore, results suggest using an intermediate (20, 30) to large number of clusters 
for improving the \emph{k}-means clustering performances, with a preference for the Short encoding.

\subsection{Qualitative Cluster Analysis}

As an additional way to validate the clustering results, we seek to visualize the clusters, and compare them to the natural layout of the reduced data.

\begin{figure}[ht]
  \begin{subfigure}{\linewidth}
  \includegraphics[width=0.99\linewidth]{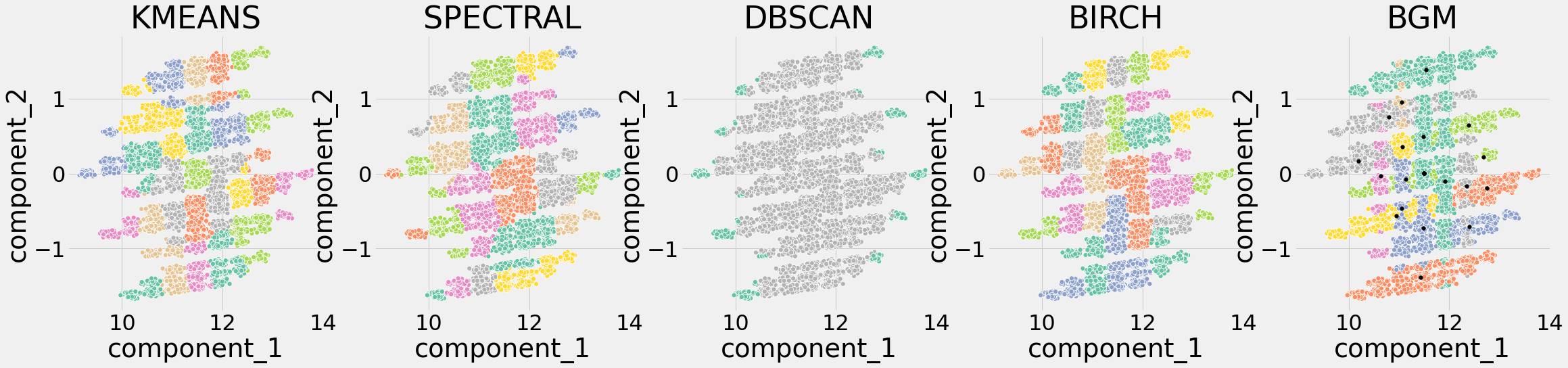}\hfill%
  \caption{}
  \end{subfigure}\par\medskip
  \begin{subfigure}{\linewidth}
  \includegraphics[width=.99\linewidth]{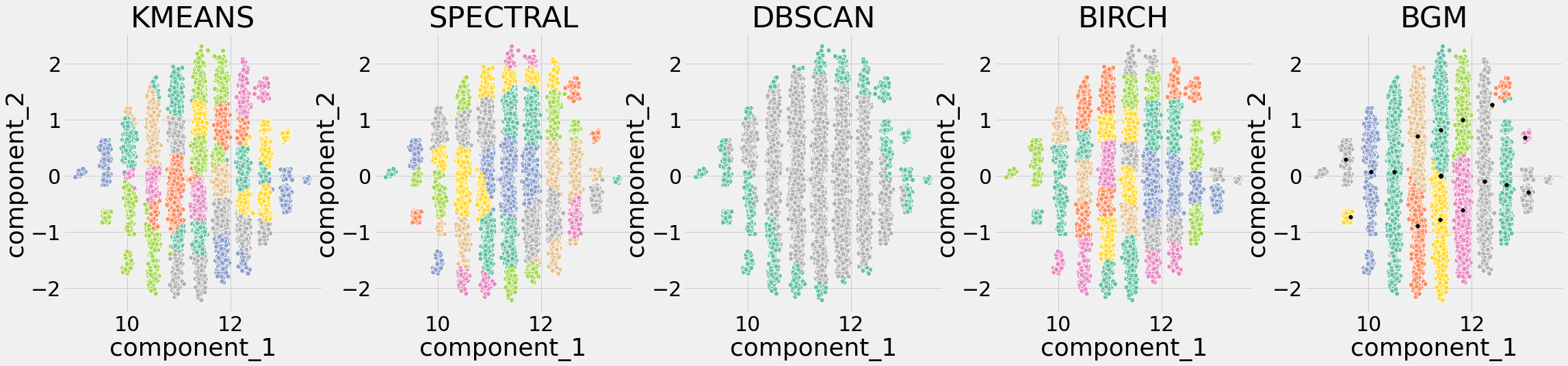}\hfill
  \caption{}
  \end{subfigure}
  \caption{Qualitative analysis of clustering for both feature representations. 
  Here we use Truncated SVD for dimension reduction, and $N=2$ components.}
  \label{fig:qualitative}
\end{figure}

Figure~\ref{fig:qualitative} depicts visual clustering results for five algorithms: \emph{k}-means, spectral clustering, DBSCAN, Birch, and BGM. All input features were reduced to two (2) components using Truncated SVD.
Plots (a) and (b) display results using the Short and the Long encoding, respectively.

When using the Short encoding (Figure~\ref{fig:qualitative}a), the clusters seem to have natural horizontal to diagonal (45 degree) layout.
This layout is not well captured by the evaluated algorithms.
The BGM seems to provide the most satisfying results, despite little calibration.

When using the Long encoding (Figure~\ref{fig:qualitative}b),
clusters naturally layout in well separated vertical columns.
This is also best captured by BGM. 

Overall, results suggest using BGM for robust clustering on both feature representations.

\subsection{Initialization Benchmark}
\label{subsec:init-benchmark}


In order to assess the quality of the centroids extracted, 
we use them for initializing the baselines algorithms GA, ($\mu + \lambda$)EA, and \textit{Aging Evolution}.

\begin{figure}[ht]
\centering
  \includegraphics[width=0.99\linewidth]{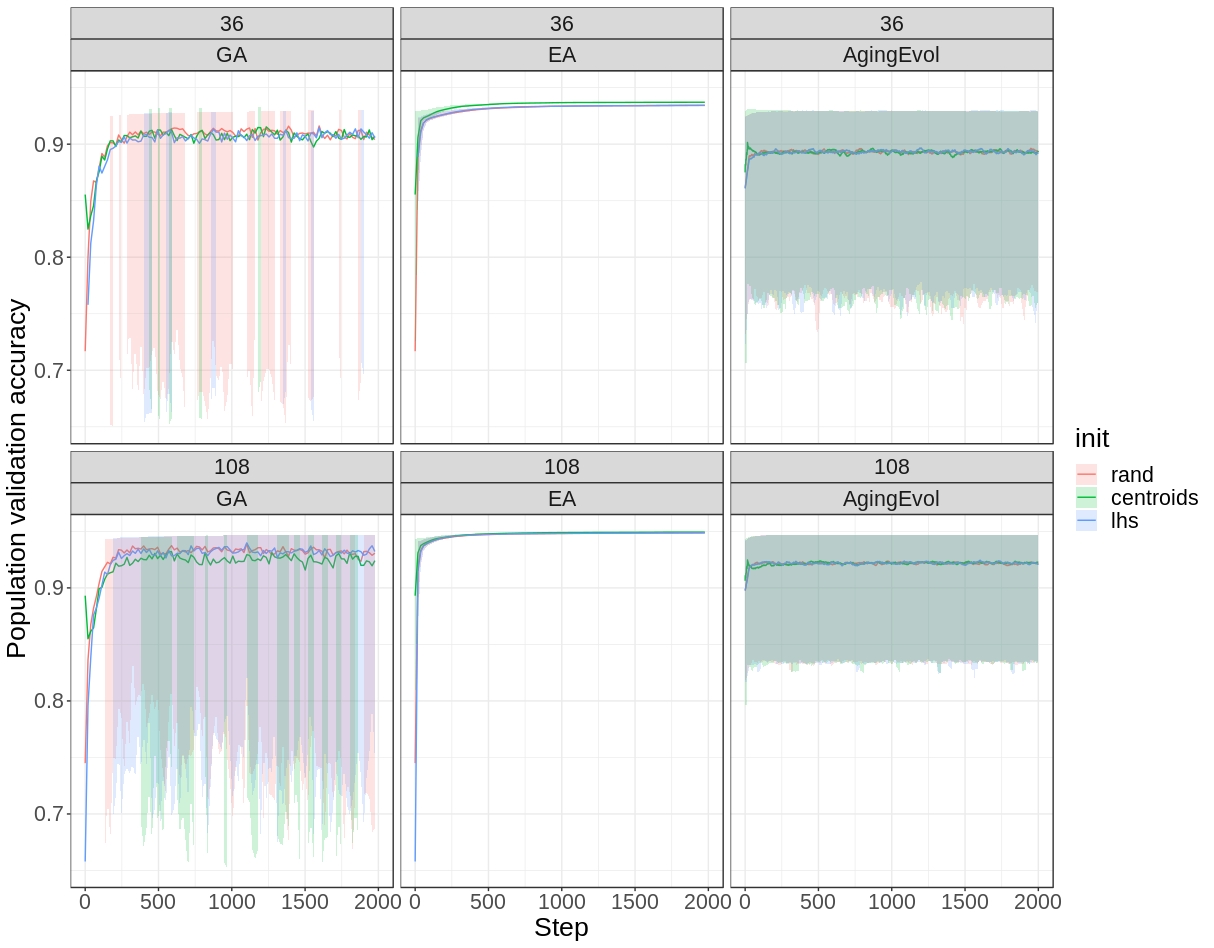}\hfill
  \caption{
  Performances in validation of various NAS algorithms, when clustering with the Short Encoding.
  }
  \label{fig:convergence-short-enc}
\end{figure}

Figure~\ref{fig:convergence-short-enc} shows performance in validation for the three search baselines. 
The color red stands for the random sampling initialization (\emph{rand}), blue for LHS, and green for centroids (i.e., our approach).
From left to right, the GA, EA, and Aging Evolution results are plotted.
The top row corresponds to 36 epochs of training, and the bottom one to 108 epochs.
In all cases, each algorithm is executed 100 independent times. 
Each plot provides with the mean fitness of the current population (bold), complemented with the range of fitness (min/max).
The centroids are initialized considering the Short encoding and BGM previous results.
The population size is set to 19 (i.e., the number of centroids) in all cases. 

Regarding GA (Algorithm~\ref{algo:ga}), \textit{pop\_size}=19, \textit{max\_evaluations}=1995 (i.e., population size 19, evolved 104 generations), \textit{mut\_p}=0.2, \textit{cx\_p}=0.5, and \textit{ind\_pb}=0.05.
Regarding EA (Algorithm~\ref{algo:mulambda}), $\mu=\lambda$=19, \textit{mut\_p}=0.8, and \textit{ind\_pb}=0.1.
Regarding Aging Evolution (Algorithm~\ref{algo:aging-evol}), it is run with a tournament size~$k=10$.

For all three baselines, we observe that the centroid-based initialization provides with the highest initial mean population fitness.
On the other hand, both LHS and random sampling-based initialization provide with a very low initial mean fitness (up to 20 percentage points of difference).
The EA takes the best advantage of this improved initial population: 
It converges faster and has long-term improvements over an initialization with random sampling or LHS.
Both GA and Aging Evolution fail to benefit from such improvements as their mean population fitness plummets after a few iterations,
and reaches similar values to those of the alternative initialization techniques (rand or LHS).
This is observed when searching either after 36 or 108 epochs of training.

\begin{figure}
\centering
  \includegraphics[width=0.9\linewidth]{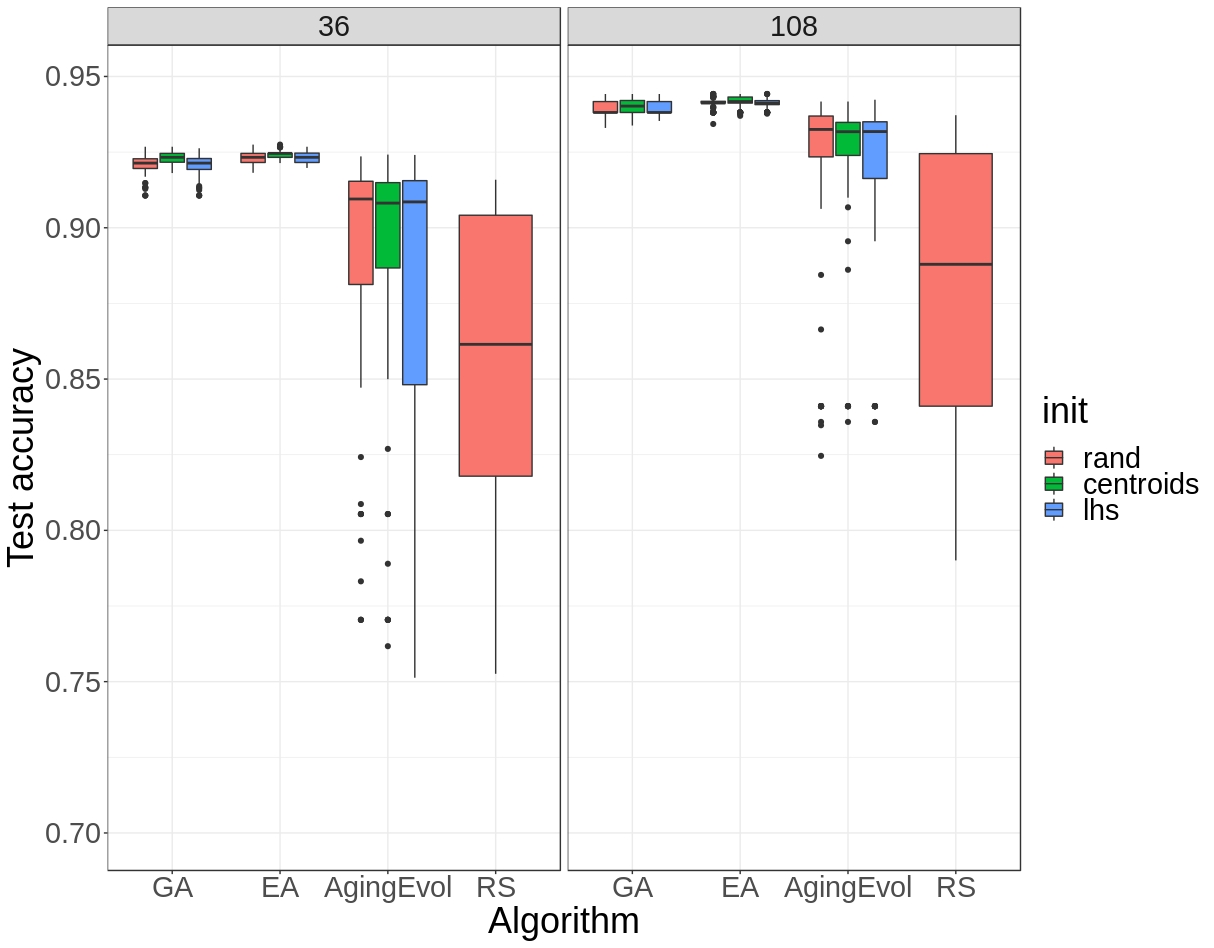}\hfill
  \caption{
  Benchmark of NAS algorithm performances after 2000 iterations. 
  The search is performed either when training solutions for 36 or 108 epochs.
  The data-driven initialization techniques involve the Short encoding.
  }
  \label{fig:test-boxplot-short-encoding}
\end{figure}

Figure~\ref{fig:test-boxplot-short-encoding} summarizes the benchmark provided in Figure~\ref{fig:convergence-short-enc}. 
In particular, it provides boxplots of performance in test for the best found solutions (100 runs) after 2000 search evaluations.
It also complements the three baseline algorithms, i.e., GA, EA, and Aging Evolution, with random search (RS).
The plot on the left corresponds to 36 epochs of training (i.e., the evaluation of the solutions), and the right one to 108 epochs.

Overall, performances in test after deployment (2000 evaluations) are similar to those in validation.
Indeed, the ranking is preserved: The EA reaches the highest mean fitness, for all initialization settings.
EA and GA have very narrow fitness distributions, while Aging Evolution has a more spread one.
All the baselines improve over Random Search.
For GA and EA, centroids help reach higher mean test fitness over other initialization techniques.

To complement these results, we performed a Wilcoxon rank-sum test.
For GA and when selecting after 36 epochs of training, the $p$-value for the centroid-based initialization versus random sampling is $4.093\cdot10^{-7}$.
Versus LHS, it is $2.324\cdot10^{-7}$.
When selecting after 108 epochs, it is $0.69$ versus random sampling, and  $0.039$ versus LHS.
For EA and when selecting after 36 epochs of training, the $p$-value for the centroid-based initialization versus random sampling is $5.611\cdot10^{-8}$.
Versus LHS, it is $3.767\cdot10^{-6}$.
When selecting after 108 epochs, it is  $0.006$ versus random sampling, and  $0.006$ versus LHS.
Thus, the centroid-based initialization significantly improves over LHS and random sampling, for both EA and GA when selecting after 36 epochs of training. For EA, it improves significantly over LHS and random sampling in all training budgets.

\begin{figure}
\centering
  \includegraphics[width=0.99\linewidth]{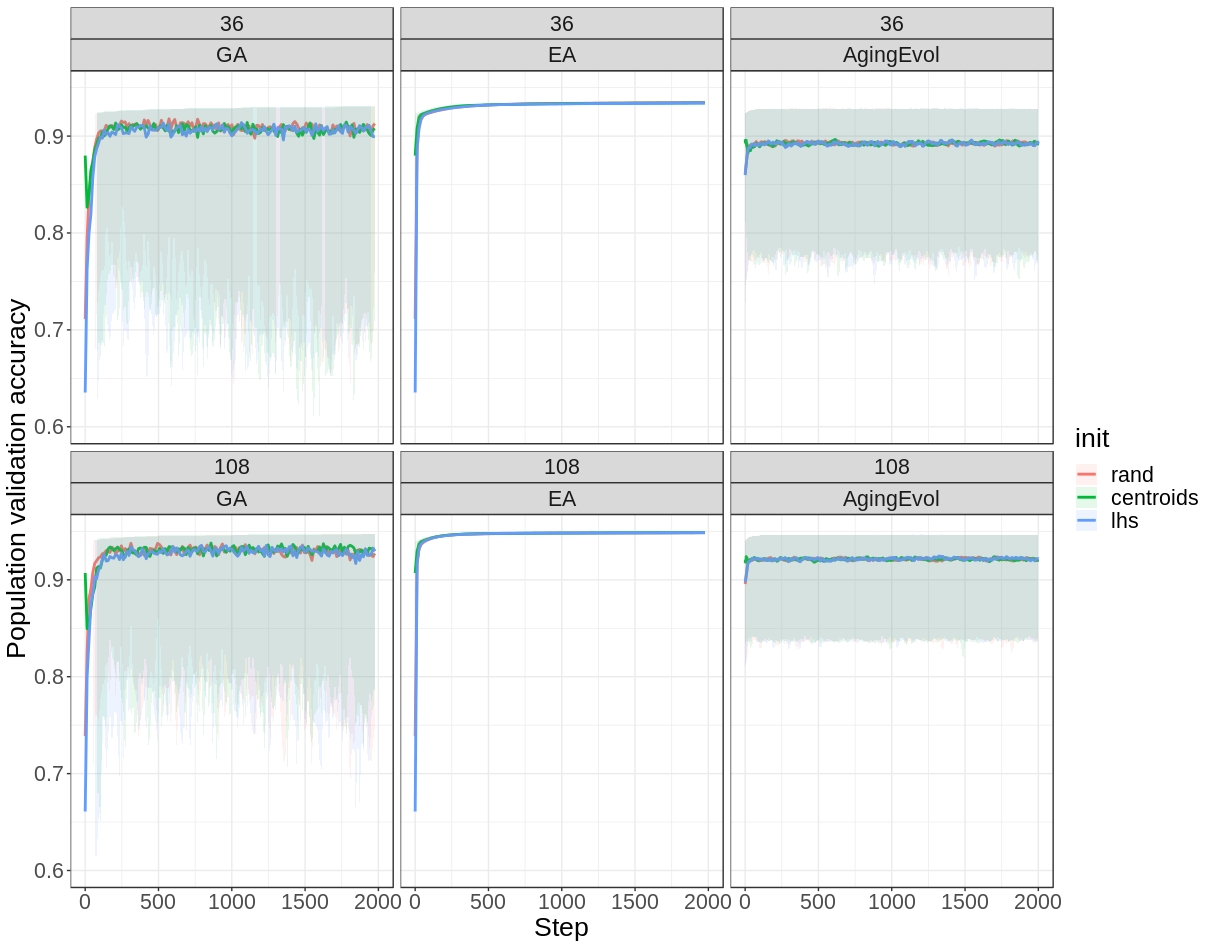}\hfill
  \caption{
  Performances in validation of various NAS algorithms, when clustering with the Long encoding.
  }
  \label{fig:convergence-long-enc}
\end{figure}

Figure~\ref{fig:convergence-long-enc} depicts the results for Long encoding benchmark. In all cases the \textit{pop\_size}=13 (=$\mu=\lambda$).
Accordingly, the number of evaluations is set to 1989 (153 generations). 

Similarly, centroids obtained considering the Long encoding enable all baseline algorithms 
to have an improved initial mean population fitness.  
Also, EA is the best at taking advantage of this initialization (centroids), with an improved convergence, up until 500 to 1000 evaluations.

\begin{figure}
\centering
  \includegraphics[width=0.85\linewidth]{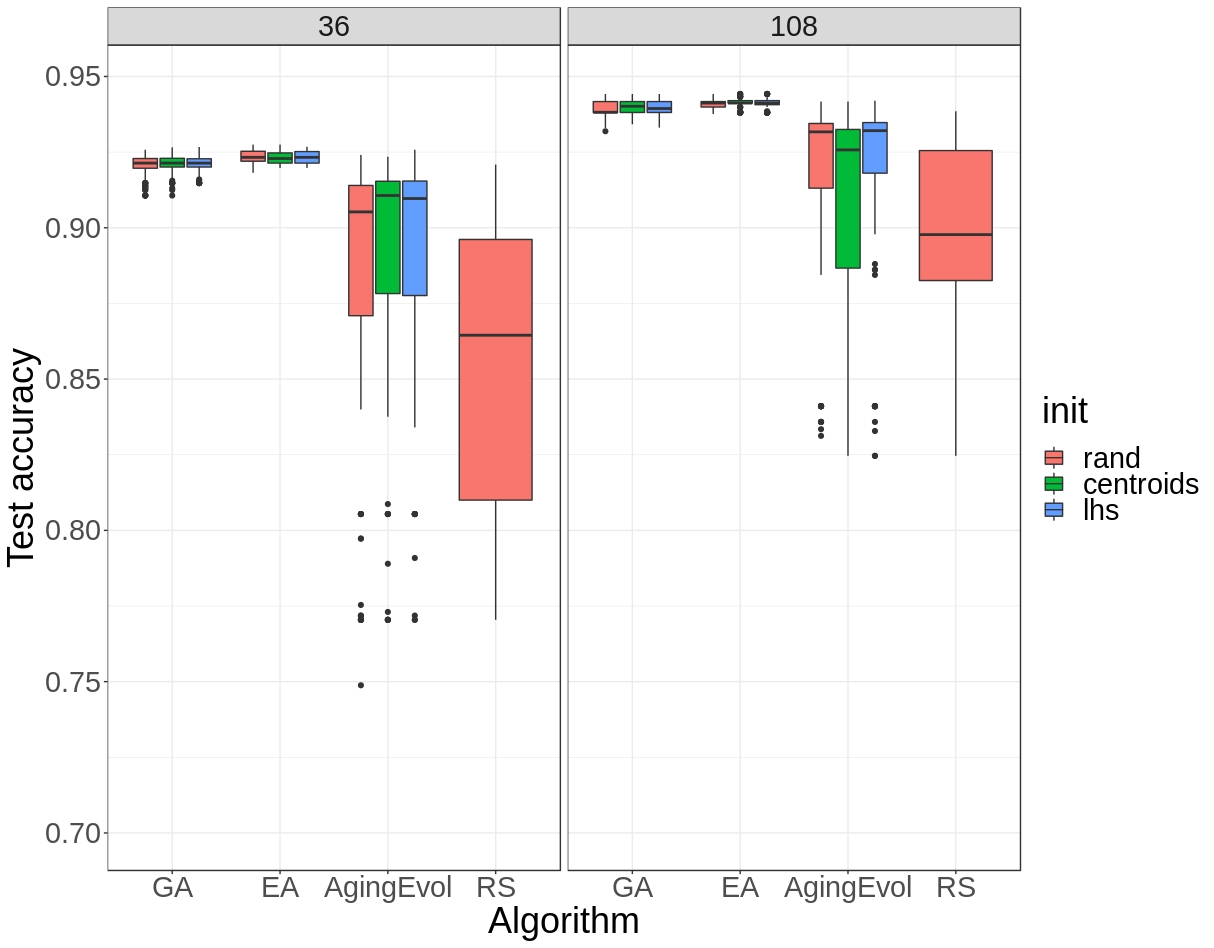}\hfill
  \caption{
  Benchmark of NAS algorithm performances after 2000 iterations.
  The search is performed either when training solutions for 36 or 108 epochs.
  The data-driven initialization techniques involve the Long encoding.
  }
  \label{fig:test-boxplot-long-encoding}
\end{figure}

Figure~\ref{fig:test-boxplot-long-encoding} summarizes the benchmark provided in Figure~\ref{fig:convergence-long-enc} with performances in test,
in the same fashion as Figure~\ref{fig:test-boxplot-short-encoding}.

Results of performance in test after deployment are similar to those obtained considering the Short encoding for finding the initial population.
However, the centroids do not provide with improvements to the final performance of the baseline algorithm.
Also, we notice that the centroid-based initialization worsen the distribution of fitness of solutions found by Aging Evolution (i.e., larger variance).

To summarize, 
the centroids extracted from a fitness-based clustering of the search space 
seem to be a promising strategy to initialize a population-based search algorithms. 
We observe improved convergence and long-term performances of EA with a centroid-based initialization, over LHS and rand, 
when considering the Short encoding. In the case of searching with only 36 epochs of budget, it also helps final test performances for GA (Short encoding). 



The limited improvements when clustering with the Long encoding 
might be explained by the fact that the baselines (EA, GA, and \textit{Aging Evolution}) are
deployed on models using the Short encoding. 
Note that experiments using the Long encoding were discarded 
because of the increased complexity for the search procedure.  
Future work might explore this option,
as it could help better exploit the extracted population.

\subsection{Visualization of the solutions found}

Last but not least, 
we look to gain insights into the solutions found by the algorithms deployed in 
Section~\ref{subsec:init-benchmark}.

\begin{figure}
      \begin{subfigure}{\linewidth}
          \includegraphics[width=0.99\linewidth]{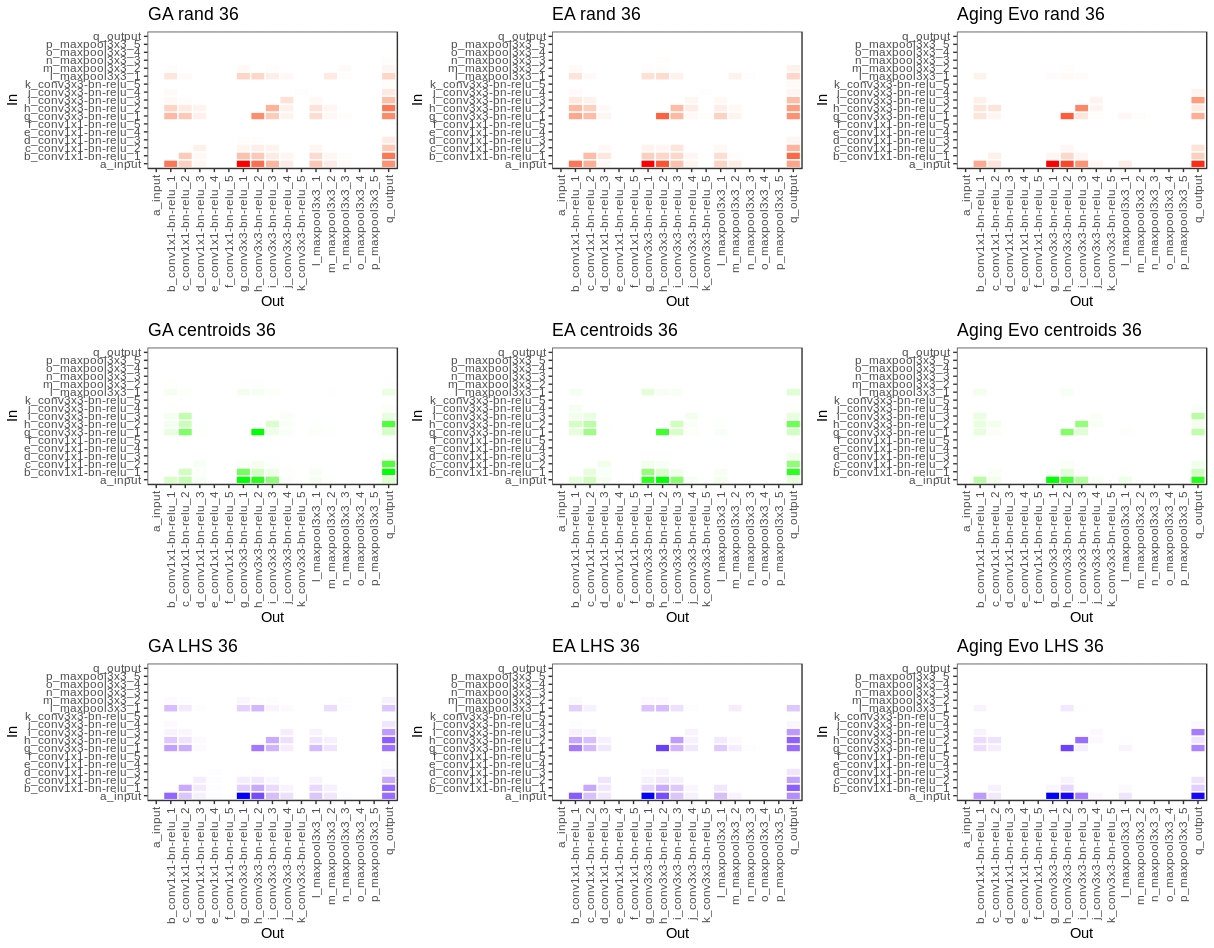}\hfill
          \caption{}
          \label{subfig:matrix-solutions-short-a}
      \end{subfigure}\par\medskip
      \begin{subfigure}{\linewidth}
          \includegraphics[width=0.99\linewidth]{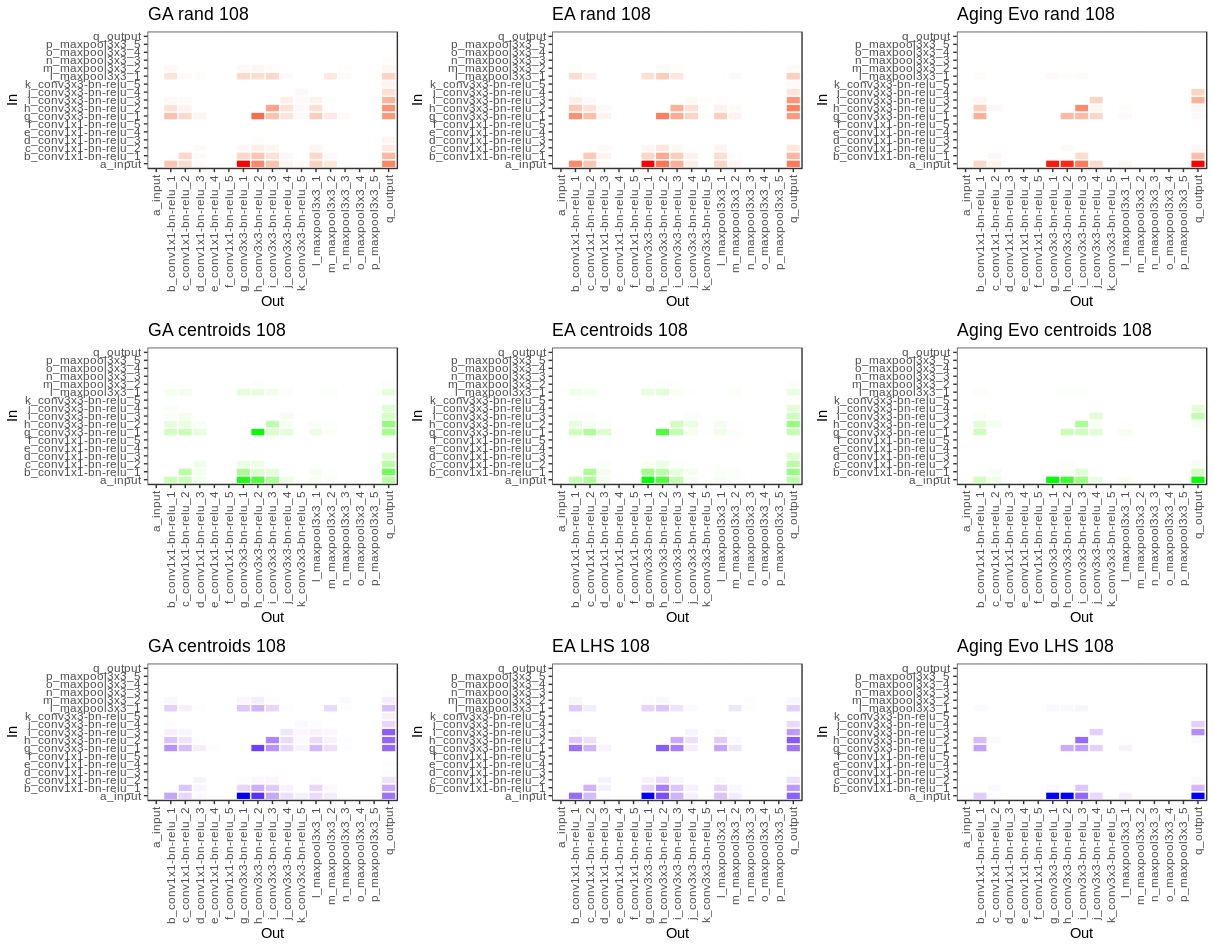}\hfill
          \caption{}
          \label{subfig:matrix-solutions-short-b}
      \end{subfigure}
  \caption{Visualization of solutions found (N=100) considering the Short Encoding.}
  \label{fig:matrix-solutions-short}
\end{figure}

Figure~\ref{fig:matrix-solutions-short} provides a visualization of solutions found (100 independent runs) by the search baselines, 
for all initialization settings, considering the Short encoding.
More precisely, it shows the frequency of connections on the adjacency matrix (100 solutions), for each baseline.
The darker, the higher the frequency. 
Figures~\ref{subfig:matrix-solutions-short-a} and~\ref{subfig:matrix-solutions-short-b} show results when searching respectively after 36 or 108 epochs of training.  
From left to right appear results for GA, EA and Aging Evolution.
From top to bottom appear results using as initialization random sampling (rand), centroids, and LHS. 

\begin{figure}
      \begin{subfigure}{\linewidth}
      \includegraphics[width=0.99\linewidth]{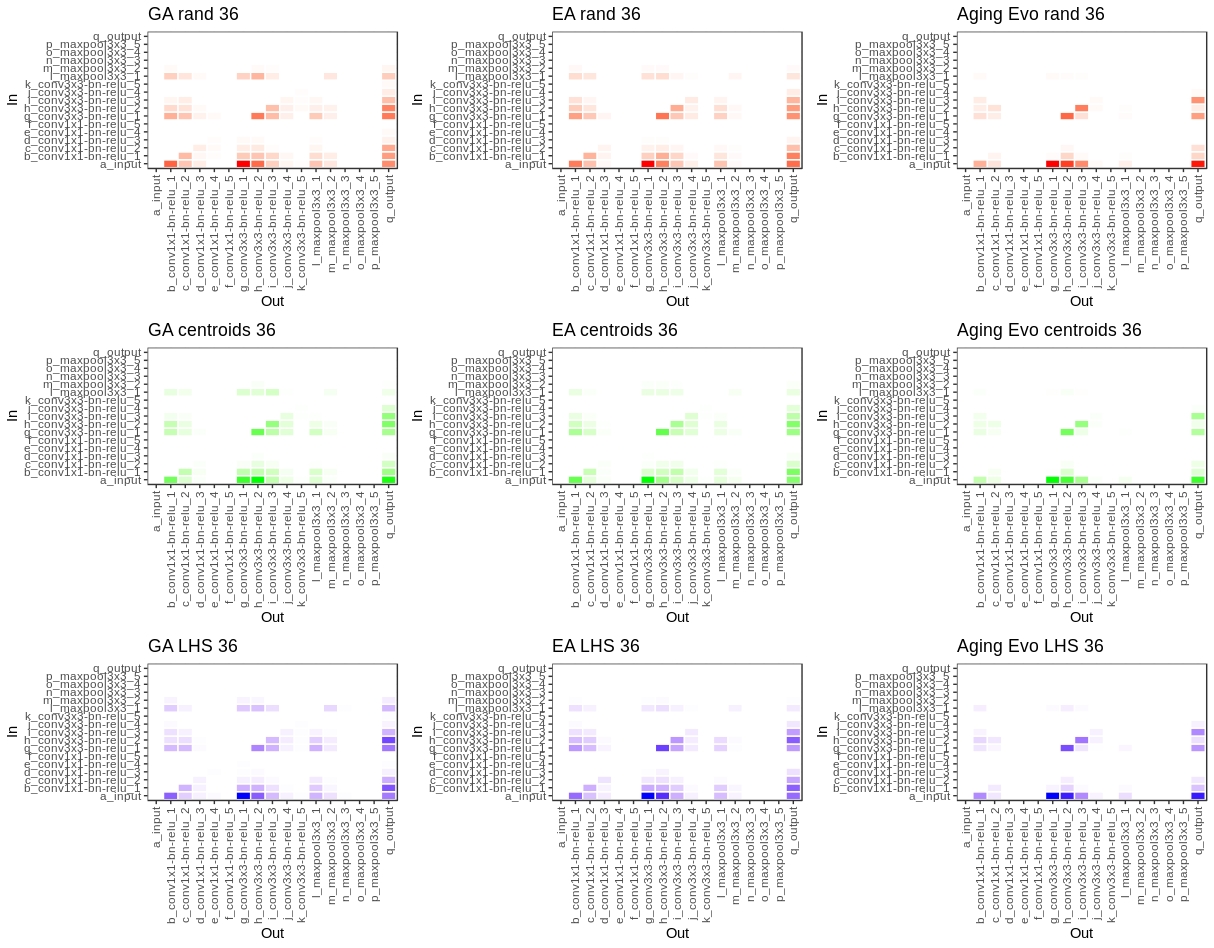}\hfill
      \caption{}
      \end{subfigure}\par\medskip
  \begin{subfigure}{\linewidth}
      \includegraphics[width=0.99\linewidth]{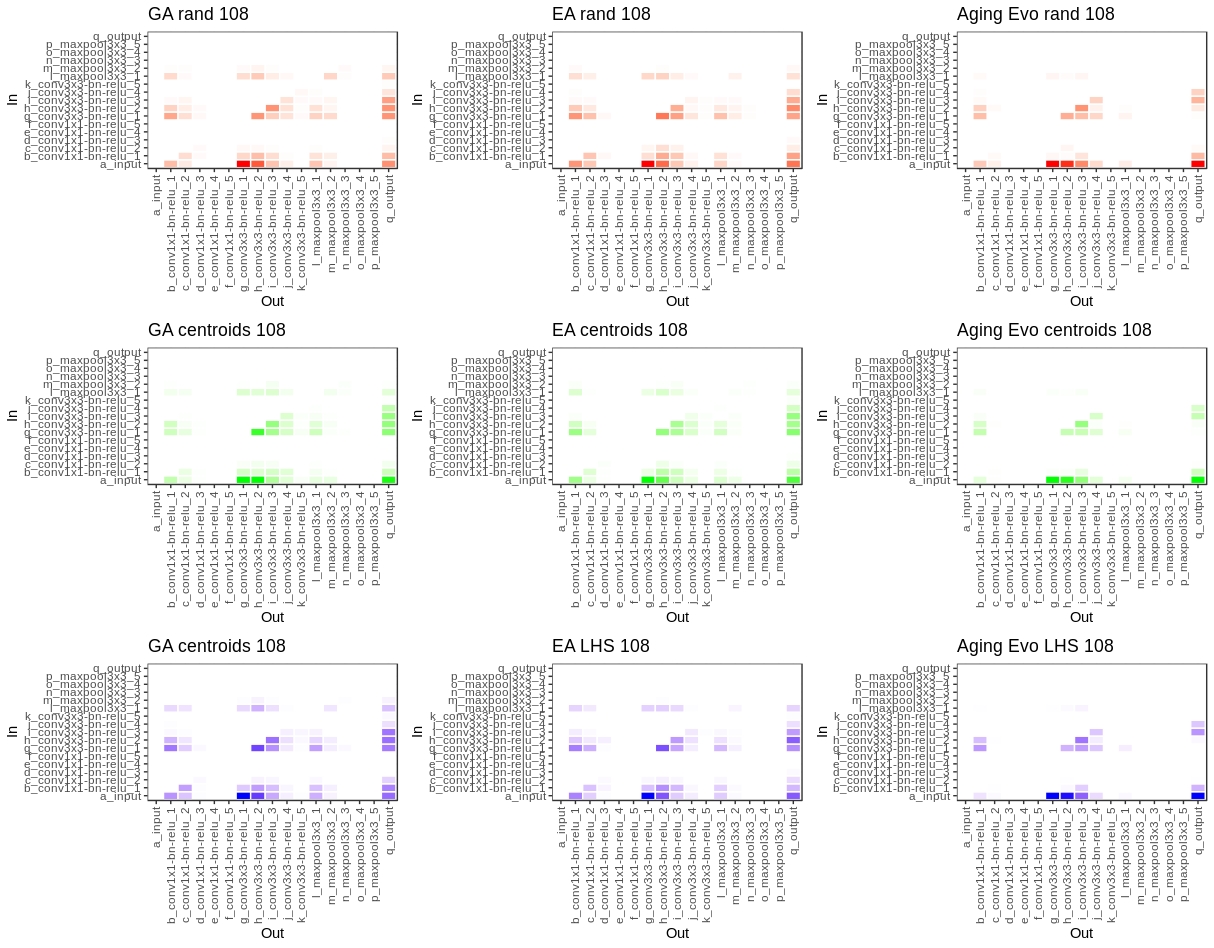}\hfill
      \caption{}
      \end{subfigure}
      \caption{Visualization of solutions found (N=100) considering the Long Encoding.}
  \label{fig:matrix-solutions-long}
\end{figure}

Figure~\ref{fig:matrix-solutions-long} provides the same visualization of solutions found but considering the Long encoding.

Overall, the connections gathered from the solutions found after 36 epochs of training differ 
from those found after 108 epochs.
In the first case, the \emph{activations} on the adjacency matrices have clusters that are more restricted, 
as opposed to the more widespread and larger clusters obtained when searching after 108 epochs of training.

Besides, we also observe a difference in the output based on the algorithm used to find the solutions.
EA and GA provide solutions whose connections are overall similar, in the form of widespread clusters.
On the other had, the Aging Evolution has patterns of connections in its cells that are regrouped and in slightly smaller cluster.

Furthermore, we analyzed the solutions based on the initialization technique used when deploying search. 
Across all settings, it appears to be more diverse solutions (on average more activated cell in adjacency matrices) obtained via LHS and random sampling, 
than for a centroid-based initialization.

To summarize, the longer the training allowed when selecting models, the more diverse are the solutions retrieved.
Also, EA and GA tend to find more diverse solutions than Aging Evolution.
When it comes to initialization, the centroids-based approached results in solutions that are more similar to each other, 
with matrices of adjacency that are less activated.

We find that the patterns highlighted in this section correlate with the findings of authors in~\cite{traore2021fitness}.
In the study, the authors show that on the search space of NASBench-101,  
the longer the training the more narrow the fitness distribution with most solutions having close to the top fitness after 108 epochs of training.
They also showed that the fitness landscape becomes flat, with many local optima.
Therefore, when searching with a training budget of 108 epochs and a fixed number of iterations, 
a search algorithm is likely to retrieve more diverse solutions than after 36 epochs, since most of them satisfy the criterion of high fitness. 

When it comes to the differences based on the algorithm to be used, this could be explained both the very rugged landscape (many local maxima) 
and the nature of the algorithms. 
Indeed, as Aging Evolution provides with non-diverse sets of solutions, which could be explained by it being stuck in local maxima and 
not diversifying enough, i.e., discarding \emph{old} solutions. 

Regarding the centroids, Section~\ref{subsec:init-benchmark} already shows that they consist of an initial population of particularly high average fitness, with little variance. This could be explained by the centroids being potential local maxima of high fitness and very diverse nature, since coming from distinct clusters.



\section{Conclusion}\label{sec:conclusion}


In this study, we seek to gain insights about a search space of image classification models
in order to improve the performance of NAS algorithms.
More precisely, we want to know if the convergence of a search strategy could be improved 
using a data-driven initialization technique exploiting the search space.

For this purpose, we propose a two-step approach to improve the performances of a NAS search strategy.
First, we perform a clustering analysis of the search space, involving a sequence of sub-tasks.
It summarizes as follows: We sample models from a search space, reduce their dimension, perform a clustering.  
After a careful tuning of the clustering pipeline (number of dimensions, clusters, etc.), 
we select the algorithm providing the best qualitative and quantitative results.
Second, we extract and use the centroids as an initial population to a search strategy.

We validate our proposal by initializing three (3) 
evolutionary algorithms, namely a genetic algorithm (GA), an evolutionary algorithm (EA), and Aging Evolution (AE), 
and benchmark our data-driven initialization
method against conventional initialization baselines, 
i.e., random initialization and Latin Hypercube Sampling (LHS). 
To test the algorithms, we query the dataset of NAS-Bench-101, providing with a search space of image classifiers and their fitness evaluation on CIFAR-10.
Our results show that centroids extracted using BGM for clustering are a promising approach to 
initialize a population-based algorithm.
In the scenario of selecting models trained only 36 epochs, this approach used with GA shows significant long-term improvements (after 2000 iterations, in test) over random initialization and LHS, when using a Short encoding.
When used with EA, it shows faster convergence (in validation) and significant long-term improvements 
over random initialization and LHS, when using a Short encoding and for all training budgets. 
Additional investigations on the distributions of the solutions found by the algorithms suggest 
that centroids enable retrieving local optima (maxima) of high fitness and similar configurations.   

As future work, we propose to investigate performances of this approach when selecting models on the Long Encoding. We also propose to study in depth the obtained clusters to gain more insights on obtained performances. One might also explore the benefits of such data-driven initialization method on other families of algorithms (Bayesian optimization, local search, etc.).

\backmatter

%
%
%

\bmhead{Acknowledgments}

Authors acknowledge support by the European Research Council (ERC) under the European Union's Horizon 2020 research and innovation program (grant agreement No. [ERC-2016-StG-714087], Acronym: \textit{So2Sat}), by the Helmholtz Association
through the Framework of Helmholtz AI [grant  number:  ZT-I-PF-5-01] - Local Unit ``Munich Unit @Aeronautics, Space and Transport (MASTr)'' and Helmholtz Excellent Professorship ``Data Science in Earth Observation - Big Data Fusion for Urban Research''(W2-W3-100),  by the German Federal Ministry of Education and Research (BMBF) in the framework of the international future AI lab "AI4EO -- Artificial Intelligence for Earth Observation: Reasoning, Uncertainties, Ethics and Beyond" (Grant number: 01DD20001), the grant DeToL, and a DAAD Research fellowship.

\section*{Declarations}


\begin{itemize}
\item Funding: the funding is stated in Acknowledgment section. 
\item Conflict of interest/Competing interests: the authors have no relevant financial or non-financial interests to disclose.
\item Ethics approval: Not applicable. 
\item Consent to participate: Not applicable.
\item Consent for publication: All authors have checked the manuscript and have agreed to the submission.
\item Availability of data: https://github.com/kalifou/data-driven-initialization-to-search.
\item Code availability: https://github.com/kalifou/data-driven-initialization-to-search.
\end{itemize}







\begin{appendices}






\end{appendices}


\bibliography{sn-bibliography}


\end{document}